\documentclass[fleqn,10pt]{wlscirep}
\usepackage[utf8]{inputenc}
\usepackage[T1]{fontenc}

\newcommand{\new}[1]{\textcolor{black}{#1}}

\title{\new{Model architecture can transform catastrophic forgetting into positive transfer}}

\author[1,*]{Miguel Ruiz-Garcia}
\affil[1]{Department of Mathematics, Universidad Carlos III de Madrid, 28911 Legan\'es, Spain}

\affil[*]{miguel.ruiz.garcia@uc3m.es}


\begin{abstract}
The work of McCloskey and Cohen popularized the concept of catastrophic interference. They used a neural network that tried to learn addition using two groups of examples as two different tasks. In their case, learning the second task rapidly deteriorated the acquired knowledge about the previous one. \new{We hypothesize that this could be a symptom of a fundamental problem: addition is an algorithmic task that should not be learned through pattern recognition. Therefore, other model architectures better suited for this task would avoid catastrophic forgetting.} We use a neural network with a different architecture that can be trained to recover the correct algorithm for the addition of binary numbers. \new{This neural network includes conditional clauses that are naturally treated within the back-propagation algorithm.} We test it in the setting proposed by McCloskey and Cohen and training on random additions one by one. The neural network not only does not suffer from catastrophic forgetting but it improves its predictive power on unseen pairs of numbers as training progresses. \new{We also show that this is a robust effect, also present when averaging many simulations}. This work emphasizes the importance that neural network architecture has for the emergence of catastrophic forgetting and introduces a neural network that is able to learn an algorithm.
\end{abstract}

\begin{document}

\flushbottom
\maketitle
%
%
\thispagestyle{empty}

\section*{Introduction}

Catastrophic forgetting or \new{catastrophic} interference is today a central paradigm of modern machine learning. It was introduced in 1989 by McCloskey and Cohen \cite{mccloskey1989catastrophic} when they systematically showed how learning new information by a fully connected neural network leads to the rapid disruption of old knowledge. This work was followed by many others that confirmed this phenomenology using a myriad of different models and tasks \cite{ratcliff1990connectionist,lewandowsky1995catastrophic,french1999catastrophic,goodfellow2013empirical,srivastava2013compete,nguyen2019toward,mirzadeh2020understanding,lee2021continual}.

In this context, a neural network (NN) \new{represents} a nonlinear mapping between input and output spaces. This nonlinear mapping depends on many parameters (the weights of the NN) that can be tuned to achieve the desired behavior, e.g. that a specific image (input) maps to the correct class in the output. In supervised classification tasks, many labeled examples are used to build a loss function that quantifies the performance of the NN. Training the NN consists on \new{minimizing} this loss function with respect to the weights, so that the system moves in parameter space to a region where it correctly classifies most of the training data. In the classic work by McCloskey and Cohen they used a NN that took as input two numbers and output another one. They used this architecture to learn two different tasks: the NN was first trained to output the correct answer to the ``ones addition facts'' ($1+1$ through $9+1$ and $1+1$ through $1+9$). Once the network learned the first task, they trained it on the ``twos addition facts''  ($1+2$ through $9+2$ and $2+1$ through $2+9$). Although they changed many components of their setup, they always found that training on the second task erased any knowledge about the previous one. In other words, when the system tried to \new{minimize} the loss function corresponding to the second task, it moved to a region of parameter space where it was not able to solve the first task. Inspired by human and animal learning \cite{mcclelland1995there,braun2001dynamic,barnett2002and,yang2014sleep,cichon2015branch,musslick2017multitasking,flesch2018comparing} or even by the behavior of physical systems \cite{pine2005chaos,keim2011generic,keim2014mechanical,hexner2020periodic,sachdeva2020tuning,stern2020supervised,stern2020continual,dillavou2021demonstration}, modern approaches to this problem try to constrain the change of the parameters during training of the second task to avoid forgetting the first one \cite{rusu2016progressive,kirkpatrick2017overcoming,zenke2017continual,parisi2017lifelong,lopez2017gradient,shin2017continual,lee2017overcoming,riemer2018learning,de2019continual,farajtabar2020orthogonal,doan2021theoretical}.

But the definition of ``task'' can be subtle. In the case of McCloskey and Cohen, should not the addition of any two numbers be a unique task? We would definitely claim that a child has learned addition only if it is able to sum {\it any} two numbers (not only the ones already shown to them). The fact that learning a subset of additions interfered with a previously learned group of sums indicates that their NN was trying to memorize (as in pattern recognition) the results of the operations. \new{We hypothesize that, if a different NN was able to learn the algorithm of addition (instead of learning to recognize particular examples), showing new data would lead to better performance (positive transfer) instead of catastrophic interference. In the rest of this paper, we aim to create a neural network architecture that can learn the rules of addition to avoid catastrophic forgetting. We hope that this work will motivate machine learning practitioners working on different areas to consider new architectures that can transform catastrophic forgetting into positive transfer in other frameworks.
}

\section*{The mathematical model}

\subsection*{\new{ Addition of binary numbers}}

\begin{figure*}
    \centering
    \includegraphics[width=1\linewidth]{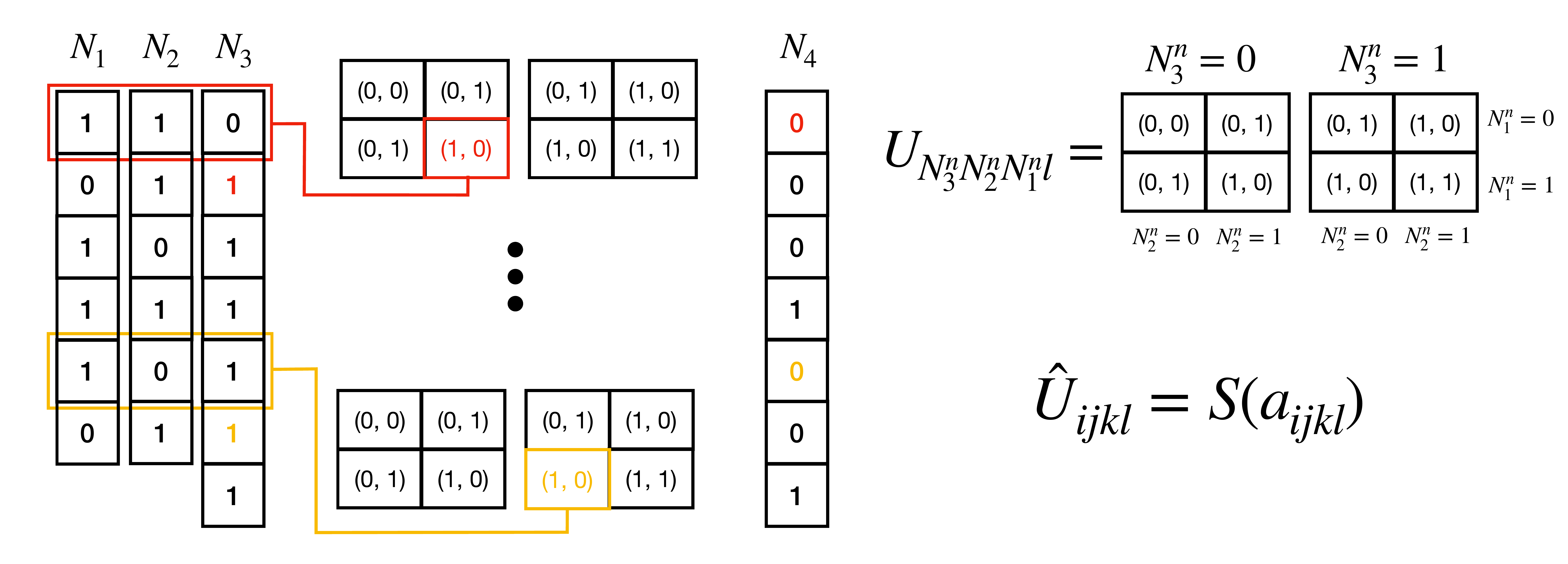}
    \caption{Addition of two binary numbers. $U_{ijkl}$ can be understood as a convolutional operator (different from the classical convolutional filters). It is applied to the input ($N_1$, $N_2$) line by line. Line nth of the input has values $N_3^n$, $N_2^n$ and $N_1^n$ that correspond to indices $ijk$ respectively. The chosen array $U_{N_3^nN_2^nN_1^nl}$ provides the output $N_3^{n+1}$ and $N_4^n$. $N_4$ corresponds to the result of $N_1+N_2$, in this case $29+43=72$. $\hat{U}_{ijkl}$ corresponds to the \new{operator used in our} NN, with parameters $a_{ijkl}$ that are learned through training. $\hat{U}_{ijkl}$ acts on $N_3^n$, $N_2^n$ and $N_1^n$ \new{analogously to $U_{ijkl}$}, although in this case $N_3^n$ can be a real number different from $0$ or $1$, in that case the output is a combination of both options, weighted with $N_3^n$. See the main text for more details.}
    \label{fig:addition}
\end{figure*}

Computers are already programmed to perform the addition of binary numbers, we would like to define a NN that can change its internal weights to find an equivalent algorithm. We can recast the algorithm of addition of binary numbers in a way that resembles the typical structure of NNs. Figure \ref{fig:addition} represents the \new{algorithm for the} addition of two binary numbers ($N_1$ and $N_2$). It outputs the correct result $N_4$ and it also \new{creates} an additional array, $N_3$, keeping track of the ones that are carried to the next step. Each of these binary numbers ($N_m$) \new{are composed of binary digits ($\{1, 0\}$)} that we notate $N_m^n$\new{, where $n$ is an index indicating the position of the binary digit.}

\new{To add two binary numbers---see Fig. \ref{fig:addition}---, the operator $U_{ijkl}$ acts secuencially on the input arrays ($N_1^n$, $N_2^n$ and $N_3^n$) and outputs two numbers at each step, $U_{N_3^nN_2^nN_1^nl}$, that correspond to $N_3^{n+1}$ and $N_4^n$, where $N_4$ is the result of the sum.} In this way, $U_{ijkl}$ acts similarly to a convolutional filter, with two main differences:
\begin{itemize}
    \item Instead of performing a weighted average of $N_1^n$, $N_2^n$ and $N_3^n$, it uses their value to choose one option (it is performing three \new{conditional} clauses).
    \item One component of the output ($U_{N_3^nN_2^nN_1^n0}$) goes into the next line of the input ($N_3^{n+1}$). There is information from the previous \new{step} that goes into the next one.
\end{itemize}

See Fig. \ref{fig:addition} for an example. We identify $N_3^n$, $N_2^n$ and $N_1^n$ as the $ijk$ indices in $U_{ijkl}$. In the first step, $N_1^0=1$, $N_2^0=1$ and $N_3^0=0$, and the output is $U_{011l}=(1,0)$, where $(1,0)$ correspond to the next empty spaces of $N_3$ and $N_4$ ($N_3^{1}$ and $N_4^0$).  \new{Iteratively applying this operator we get the correct values for $N_3$ and $N_4$.} If $N_1$ and $N_2$ have $Z$ digits, $N_3$ and $N_4$ will have $Z+1$ digits and in the last line we just identify $N_4^{Z+1}=N_3^{Z+1}$. In the example shown in Fig. \ref{fig:addition} this algorithm gets the correct answer for $29+43=72$.

\subsection*{\new{ Building an algorithmic neural network}}

\new{We can define now the architecture of our neural network. We would like to have $U_{ijkl}$ as a specific (learned) state in its parameter space. For simplicity, we define our NN using a modified operator, $\hat{U}_{ijkl}$}:
\begin{equation}
    \hat{U}_{ijkl} = S(a_{ijkl}),
\end{equation}
where $S()$ stands for the sigmoid operator,
\begin{equation}
    S(x)=\frac{1}{1+e^{-x}},
\end{equation}
and $a_{ijkl}$, $i,j,k,l \in \{0,1\}$ are $16$ parameters that have to be learned. \new{Our neural network takes the operator $\hat{U}_{ijkl}$ and applies it to the input numbers ($N_1$ and $N_2$) as shown in Fig. \ref{fig:addition} for $U_{ijkl}$. In this process the network creates two new arrays corresponding to $N_3$ and $N_4$. The predicted answer to the input addition is $N_4$, then let us use the notation $N_4 = \mathcal{F}(N_1,N_2,a_{ijkl})$ to highlight that our neural network is a nonlinear funtion of inputs $N_1$ and $N_2$ and of the parameters $a_{ijkl}$.} There is a fundamental difference with the previous case, now the digits of $N_3$ and $N_4$ will be real numbers between $0$ and $1$ and we have to decide how to apply $\hat{U}_{ijkl}$ when $N_3^n$ is different from $\{0,1\}$.

If we apply the operator $\hat{U}_{ijkl}$ to the same example shown in Fig. \ref{fig:addition}, the first line is again $N_1^0=1$, $N_2^0=1$ and $N_3^0=0$ and the output now will be $N_3^1=S(a_{0110})$ and $N_4^0=S(a_{0111})$. In the second line we find now $N_1^1=0$, $N_2^1=1$ and $N_3^1=S(a_{0110})$, since $S(a_{0110})$ is a real number between $0$ and $1$, we compute the output of this line as a combination of $\hat{U}_{010l}$ and $\hat{U}_{110l}$ in the following way:
\begin{align}
    &N_3^2 = N_3^1 \times S(a_{1100}) + (1 - N_3^1) \times S(a_{0100}) \nonumber \\
    &= S(a_{0110}) \times S(a_{1100}) + (1 - S(a_{0110})) \times S(a_{0100}),
\end{align}
\begin{align}
    &N_4^1 = N_3^1 \times S(a_{1101}) + (1 - N_3^1) \times S(a_{0101}) \nonumber \\
    &= S(a_{0110}) \times S(a_{1101}) + (1 - S(a_{0110})) \times S(a_{0101}),
\end{align}
\new{where we use the multiplicative operator, $\times$, for clarity}.
In this way, if $N_3^1=S(a_{0110})$ is exactly equal to $0$ or $1$ we recover the \new{theoretical algorithm shown in Fig. \ref{fig:addition}}, and when $S(a_{0110}) \in (0,1)$ the output is a combination of both options weighted by $S(a_{0110})$. \new{Applying this operator iteratively, the output of our neural network is computed, $N_4 = \mathcal{F}(N_1,N_2,a_{ijkl})$.}

\new{To define a learning process we create a loss function. In the simplest case, we would like to learn the addition of two specific numbers ($N_1 + N_2$), we can define the loss function as:
\begin{equation}
    \mathcal{L} = \sum_n (\tilde{N}_4^n - \mathcal{F}(N_1,N_2,a_{ijkl})^n)^2,
    \label{eq_loss}
\end{equation}
where $\tilde{N}_4$ is the correct result of the sum $N_1 + N_2$, and $\tilde{N}_4^n$ and $\mathcal{F}(N_1,N_2,a_{ijkl})^n$ are the nth binary digit of $\tilde{N}_4$ and $\mathcal{F}(N_1,N_2,a_{ijkl})$, respectively. To compute $\mathcal{F}(N_1,N_2,a_{ijkl})$ our model uses sums, multiplications and a nonlinear function, $S()$, resembling the standard convolutional filters extensively used in deep neural networks. Finally, we can differentiate $\mathcal{L}$ with respect to the parameters of the neural network ($a_{ijkl}$) using standard methods\footnote{We will publicly upload our code to GitHub at the time of acceptance of our manuscript}. Training the NN consists of changing the value of the parameters ($a_{ijkl}$) following $-\frac{\partial \mathcal{L}}{\partial a_{ijkl}}$, such that the loss function is minimized.}

\section*{Results}

\subsection*{\new{Learning the ones and twos addition facts: the problem proposed by McCloskey and Cohen}}

\begin{figure}
    \centering
    \includegraphics[width=0.9\textwidth]{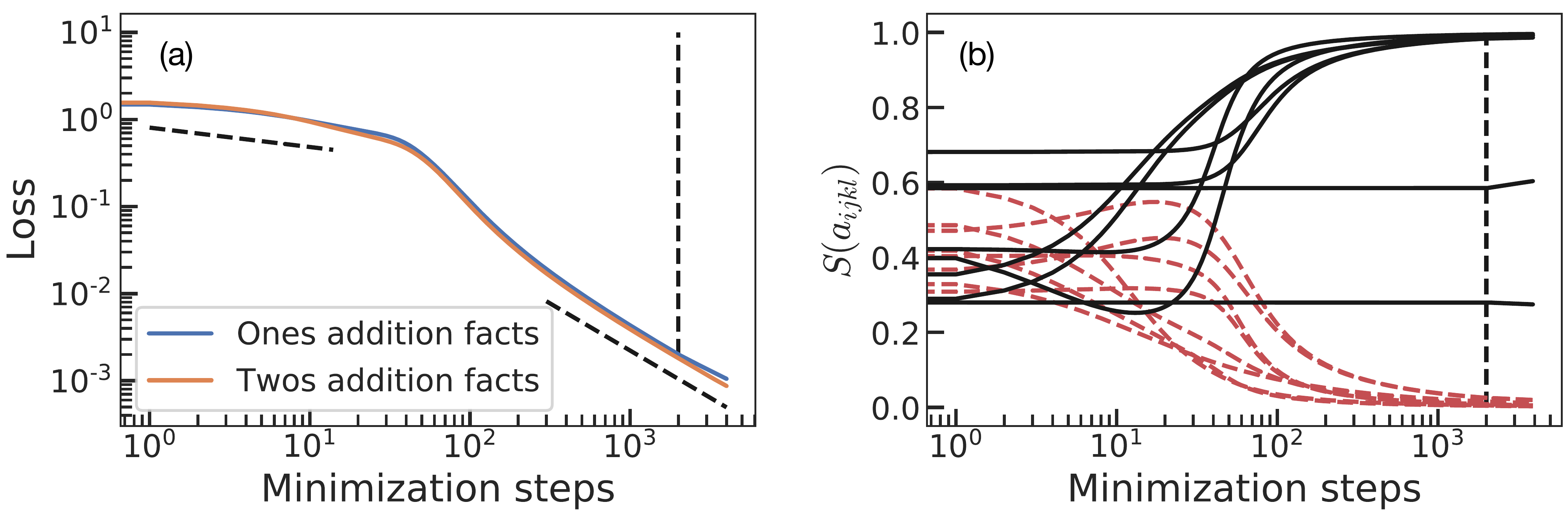}
    \caption{Training on the ones and twos addition facts. \new{Learning on each task lasts for $2000$ minimization steps (marked by the vertical black dashed lines).} Panel (a) depicts the value of the loss function for each task during training. Panel (b) shows $S(a_{ijkl})$ for the $16$ parameters of the model ($a_{ijkl}$): if learning succeeds black solid lines should tend to $1$ whereas red dashed lines should go to $0$. Two black lines stay approximately constant due to the lack of the necessary information in the dataset (see main text). \new{Learning shows two  distinct regimes with different power law behavior. The two non-vertical black dashed lines in panel (a) correspond to exponents $\sim -0.2$ and $\sim -1$.}}
    \label{fig:McCloskey}
\end{figure}

\new{In the problem proposed by McCloskey and Cohen, a neural network is trained using two tasks: the ``ones addition facts'' (all the additions of $1$ with another digit, $1+1=2$ through $9+1=10$ and $1+1=2$ through $1+9=10$) and the ``twos addition facts''  ($1+2=3$ through $9+2=11$ and $2+1=3$ through $2+9=11$). In this work, the neural network catastrophically forgets the first task when training on the second one. We aim to show that this was due to an inadequate choice of model architecture and that a different architecture, such as the one proposed in this work, will not display catastrophic forgetting when training on these tasks. }

\new{We create two datasets, one for the ones and another for the twos addition facts. We will train the model using the ones addition facts first, then we will continue training the same model using the twos addition facts. We create a loss function for each of these datasets, for each pair of numbers $N_1$ and $N_2$ we compute the correct result $N_1 + N_2 = \tilde{N}_4$ and compute their corresponding loss, using equation \eqref{eq_loss}. The total loss for each of the tasks is the average value of Eq. \eqref{eq_loss} evaluated for each pair of numbers in the task (e.g. $1+2=3$ through $9+2=11$ and $2+1=3$ through $2+9=11$). We train the network using gradient descent with learning rate equal to $1$. We train for $2000$ steps using the ``ones addition facts'' loss and for another $2000$ steps using the ``twos addition facts'' loss. In Fig. \ref{fig:McCloskey} we plot both losses during the learning process to quantify the performance of the model for both tasks. Fig. \ref{fig:McCloskey} (a) shows that the loss functions corresponding to both tasks greatly decrease when training on the ``ones addition facts'': learning the ``ones addition facts'' has a positive transfer to the ``twos addition facts''. Similarly, both loss functions keep decreasing when training on the ``twos addition facts'': there is no catastrophic forgetting and the model shows positive backward transfer from the new task to the previous one. Fig. \ref{fig:McCloskey} (b) shows the evolution of the parameters of the network. At initialization, the parameters of the network ($a_{ijkl}$) are real random numbers between $-1$ and $1$, what leads to $S(a_{ijkl})$ being randomly distributed within $1/(1+e)\sim 0.27$ and $e/(1+e)\sim 0.73$. To recover the correct addition algorithm, $U_{ijkl}$, the black continuous lines should saturate to one whereas the red dash lines should go to zero as learning progresses. Up to step $\sim 90$ in the minimization process, we observe some lines performing non-monotonic behaviors until all of them (except for two black lines) tend to the correct values, we will term this region the non-trivial learning regime. For minimization steps larger than $\sim 100$ (including when training switches to the ``twos addition facts'') learning continues smoothly and all the parameters (except two) approach their asymptotic vales, what we will term the trivial learning regime. The two black lines that do not approach $1$ correspond to $S(a_{111l})$. The reason for this behavior is that the tasks used here do not contain any addition that would require these parameters, only  used when $N_1^n=N_2^n=N_3^n=1$ (for a specific $n$), for example in $3+3$.}

\subsubsection*{\new{Learning to sum, one example at a time}}

\begin{figure}
    \centering
    \includegraphics[width=0.9\textwidth]{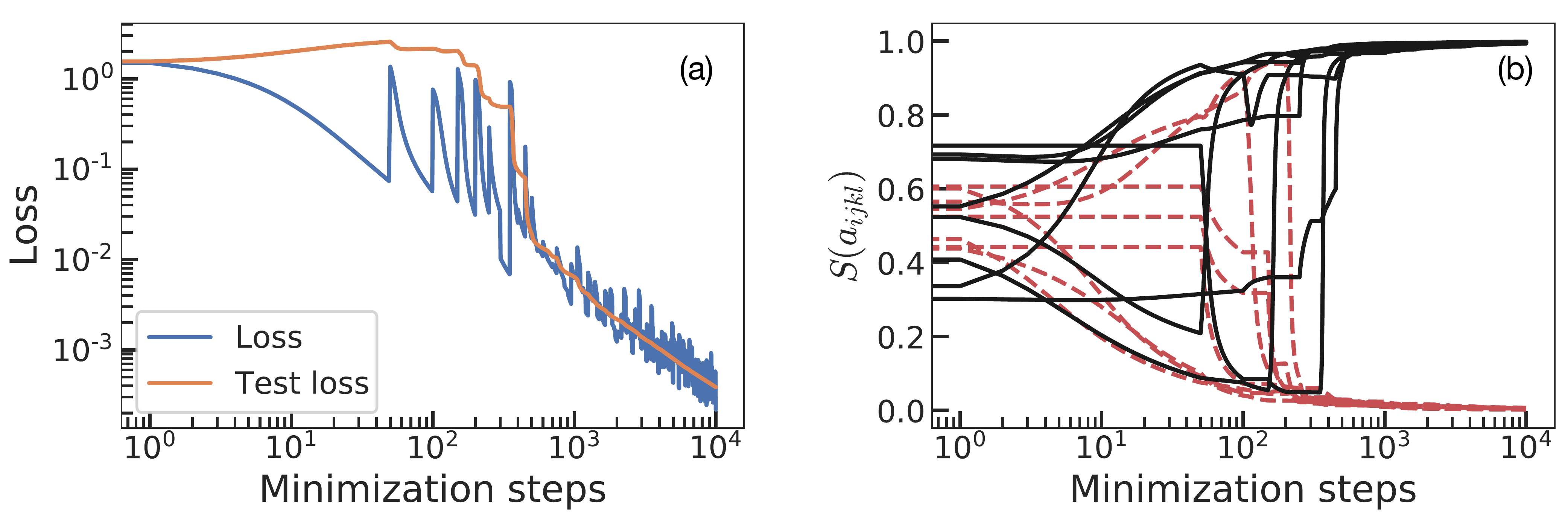}
    \caption{Learning one sample at a time. Panel (a) displays training and test loss \new{functions} during learning. We train on a new sample ($N_1+N_2=\tilde{N}_4$) chosen at random every $50$ minimization steps for $10000$ steps (a total of $200$ samples). Test loss is always computed with the same $100$ samples that are not in the training dataset. Panel (b) shows $S(a_{ijkl})$ for the $16$ parameters of the model ($a_{ijkl}$): if learning succeeds black solid lines should tend to $1$ whereas red dashed lines should go to $0$.}
    \label{fig:one_by_one}
\end{figure}

We now perform a second experiment where we train on one sample at a time. Each sample consists on two integers chosen at random ($N_1 + N_2 = \tilde{N}_4$). \new{This is more challenging than the example of the previous section. Before, the system was trained on a group of additions, the loss function was imposing more constraints on the parameters, and training led to the correct values of  $a_{ijkl}$. Now, the model is trained using one sample at a time, what gives more freedom to the parameters. We train on each sample during $50$ steps using gradient descent on Eq. \eqref{eq_loss}. Figure \ref{fig:one_by_one} shows the process  of  training the model for $200$ different samples (one after another, $10000$ steps in total). Test loss is computed every step as the average loss for $100$ samples that are not included in the training dataset.
During training we observe that the training loss (blue line in Fig. \ref{fig:one_by_one}) decreases for $50$ steps every time that a new example is shown, this indicates that the operator $\hat{U}_{ijkl}$ is changing (learning) to correctly add this particular pair of numbers. When we switch to the next sample there is a sudden increment of the loss, followed by a decrease for another $50$ steps. However, the test loss (contrary to the previous section) increases when training on the first samples, indicating the presence of overfitting: the NN is learning the addition of a pair of numbers, but the rules it is learning do not generalize to the rest of samples. From step $\sim 400$ forward, the test loss shows a steady decrease. Panel (b) of Fig. \ref{fig:one_by_one} shows the value of the parameters $S(a_{ijkl})$ at all times. Similarly to the previous section, there  is a non-trivial regime where parameters show non-monotonic behavior. In this case, they show more sudden transitions due  to  the change in training  data every $50$ steps. After step $\sim 400$ all parameters  approach their correct values leading to a trivial learning regime.}

\new{One could wonder how reproducible are the results of Fig. \ref{fig:one_by_one}, or if the learning process could get stuck in the non-trivial regime, preventing the system to reach the trivial regime and correctly learn addition. To study the robustness of this behavior we perform now $70$ different simulations in the same conditions of Fig. \ref{fig:one_by_one} (training on one example at a time for $50$ minimization steps). For each simulation we use a new random initialization and different training samples picked at random, the test dataset is the same for all the simulations. We plot the average values of the training and test loss  functions in Fig. \ref{fig:averages}. The inset shows the same data but in a semi-log scale, including shaded areas around the mean values that correspond to one standard deviation of the data. Training on each sample leads to a decreasing training loss and an increasing test loss (overfitting).  The average values show a final regime (from step $\sim5000$ forward) where all simulations collapse (note vanishing standard deviation in the inset). A fit to this regime, dashed line in Fig. \ref{fig:averages}, shows a power law behavior with an exponent $\sim -1$. Fig. \ref{fig:one_by_one} is a good representative of the behavior of most simulations contained in Fig. \ref{fig:averages}, most of then reach the trivial regime around step $\sim 1000$. However, some outliers take much longer to find the trivial learning regime, leading to  the plateau observed in the  test loss between step $\sim 1000$ and $\sim 5000$, when the last outlier reaches the trivial learning regime.}
These results prove that this model architecture shows positive transfer, changing ``tasks'' leads to better performance in previous and future tasks instead of catastrophic interference. When we stop training, our NN has learned the rules of addition, it will be able to sum {\it any} pair of numbers whether they were included in the training data or not.

\subsection*{\new{Analysis of the results}}

\begin{figure}
    \centering
    \includegraphics[width=0.55\textwidth]{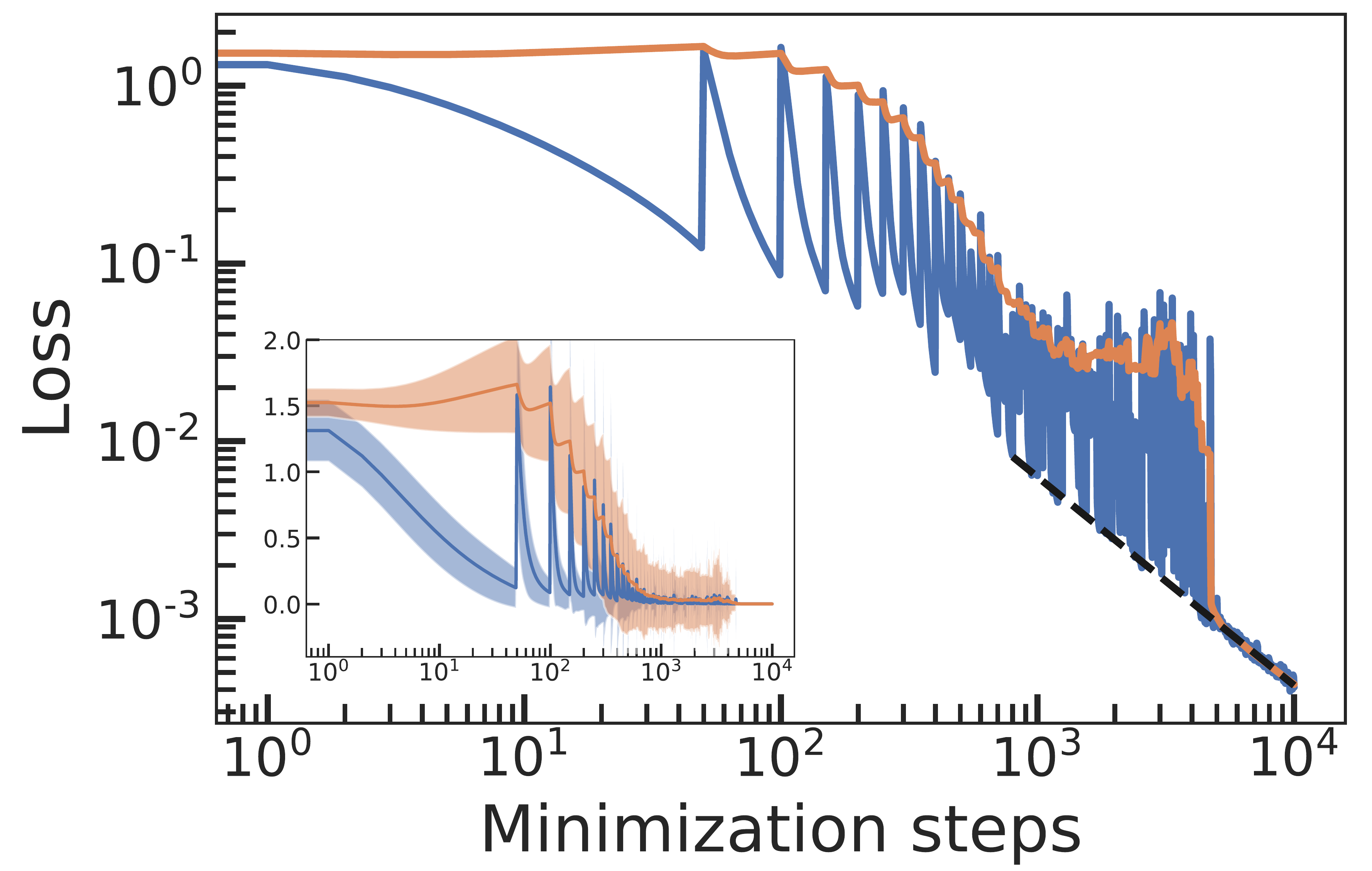}
    \caption{
    \new{Learning one sample at a time, mean values for the training and test loss. We repeat the same protocol described in Fig. \ref{fig:one_by_one} $70$ times, the model is trained using a different random example every $50$ steps. For each run we start with a different initialization and choose different training examples at random. The blue and orange lines correspond to the mean value of the training and test loss, respectively. The inset presents the same data in a semi-log scale, including a shaded area around each curve that represents one standard deviation.}}
    \label{fig:averages}
\end{figure}

\new{In the cases shown in Figs. \ref{fig:one_by_one} and \ref{fig:averages}, we use binary numbers of dimension $5$ for $N_1$ and $N_2$. The largest number considered in the additions is $11111$, $31$ in decimal form. Therefore, there are $(31*30)=930$ possible combinations to create our training and test datasets of additions. As we have seen in Figs. \ref{fig:one_by_one} and \ref{fig:averages} it was enough to take around $500$ minimization steps ($\sim 10$ different samples) to correctly learn addition and enter the trivial learning regime. After this, all the parameters start to asymptotically approach their theoretical value and the loss displays a power law decay with an exponent $\sim -1$.
}

\new{Fig. \ref{fig:McCloskey} also shows how learning in our model is divided in non-trivial and trivial learning regimes. In the former case, the evolution of the coefficients ($a_{ijkl}$) are coupled, displaying non-monotonic behavior, and it is characterized by a non-trivial exponent $0.2$, for which we do not have an analytical derivation. In the later regime, the coefficients asymptotically approach their theoretical values $a_{ijkl} \to \pm \infty $ such that $S(a_{ijkl}) \to 0, 1$. The loss function that system minimizes is a sum of terms that compare each binary digit of the correct result with the number predicted by the network, $\left(\tilde{N}_4^n-\mathcal{F}(N_1,N_2,a_{ijkl})^n\right)^2$. In this regime, the dynamics of the parameters of the network are uncoupled and all behave in an equivalent manner. For simplicity, let us study the evolution of one of the parameters that we term $a$, in this regime there are two possibilities $a \to \pm \infty$,
\begin{equation}
S(a) \to 
\begin{cases}
1 - e^{-a}, \quad \textrm{if} \quad a \to \infty,\\
e^{a}, \quad \textrm{if} \quad a \to -\infty.
\end{cases}
\end{equation}
Keeping up to leading order in $e^{-|a|}$, the terms appearing in the loss function take the following forms:
\begin{equation}
  \left(\tilde{N}_4^n-\mathcal{F}(N_1,N_2,a_{ijkl})^n\right)^2   \to 
\begin{cases}
e^{-2a}, \quad \textrm{if} \quad a \to \infty,\\
e^{2a}, \quad \textrm{if} \quad a \to -\infty.
\end{cases}
\label{eq_loss_term}
\end{equation}
All the terms included in the loss function are then proportional to the ones in equation \eqref{eq_loss_term} and we can study the evolution of one of them. Let us take the case $e^{2a}, \ a \to -\infty$. Our dynamics are discrete, but we assume that the continuous limit is a good approximation in the trivial learning regime. If we term $t$ the minimization time, the parameter $a$ evolves as
\begin{equation}
    \frac{\partial a}{\partial t}  = - \frac{\partial \mathcal{L}}{\partial a} \propto -e^{2a},
\end{equation}
we can integrate this equation as,
\begin{equation}
    \int -e^{-2a} \textrm{d}a \propto \int \textrm{d}t \implies e^{-2a} \propto t + C,
    \label{integration}
\end{equation}
where $C$ is an additive constant that would have the information about the initial condition of $a$. This constant can be neglected in the limit $t \to \infty$. Solving \eqref{integration} for $a$ we find,
\begin{equation}
    a \propto - \frac{1}{2} \ln(t),
\end{equation}
indicating that, in the trivial learning regime, the parameters $a_{ijkl}$ tend to $\pm \infty$ in a logarithmic manner. Finally, since the loss if a sum of terms proportional to the ones in equation \eqref{eq_loss_term}, using again the case $a\to -\infty$, we get,
\begin{equation}
    \mathcal{L} \propto e^{2a} \sim e^{-\ln(t)} \sim \frac{1}{t}.
    \label{eq_scaling}
\end{equation}
Equation \eqref{eq_scaling} recovers the scaling observed numerically in the trivial regime of Figs. \ref{fig:McCloskey} and \ref{fig:averages}, $\mathcal{L}(t) \sim t^{-1}$.
}

\section*{Discussion}

The NN defined in this work is able to learn to sum {\it any} two numbers when trained on a finite set of examples. But should we still consider the addition of different pairs of numbers as different tasks? It probably depends on the NN that is being used. If the NN is only able to perform some version of pattern recognition, it will not be able to extract any common rules and training on different samples will lead to catastrophic forgetting. However, if the NN has the neccesary set of tools (as shown in this work) the addition of different pairs of numbers constitute different examples of the same task, and training on all of them has a positive effect. This is similar to humans that can transfer previous knowledge when they have a deep, rather than superficial, understanding \cite{barnett2002and}.

We have defined a NN that is  algorithmically aligned \cite{xu2020neural,xu2019can} with the correct algorithm for the sum of binary numbers. This NN was able to learn the correct coefficients through gradient descent, when presenting different examples of additions. To define this NN we have created a layer that operates similarly to a traditional convolutional layer but with three important differences: 
\begin{itemize}
    \item It performs different operations based on the input \new{(equivalent to ``if'' clauses)}, instead of performing a weighted average of the input values (filter).
    \item At every step it passes information to the next line of the input ($N_3^{n+1}$).
    \item If the input ($N_3^n$) is not $0$ or $1$ the network combines both options of the corresponding ``if'' clause weighting them with $N_3^n$.
\end{itemize}
\new{The mathematical operations required to apply our model to the input data, and to build the loss function, are just sums, multiplications and the sigmoid nonlinear function, similar to standard convolutional filters. This allows us to use standard back-propagation algorithms. Since our model has $16$ parameters, we have not perfommed a systematic study of the runtimes. This could be necessary if this model were to be combined with standard deep learning layers.
}

In future work we would like to increase the complexity of this algorithmic neural network, which is able to perform different tasks depending on the input. It should be possible to combine this NN with other types of neural networks to create a complete model with the capacity to learn an algorithm at the same time that takes advantage of the power of other NNs (e.g. convolutional NNs). Hopefully, this could help ``building causal models of the world that support explanation and understanding, rather than merely solving pattern recognition problems'' \cite{lake2017building}. 
Training these complete models can lead to a non-convex optimization problem that can benefit from changing the topography of the loss function landscape \cite{ruiz2019tuning}, an effective way of doing this in machine learning is the use of dynamical loss functions \cite{ruiz2021tilting}.

In the work of McCloskey and Cohen the addition of two groups of numbers were considered as two different tasks. When training on the second task the previous knowledge was erased. However, this could have been just a symptom of using a pattern-recognition neural network for the ``detection'' of seemingly different tasks. Addition is the paradigm of algorithmic tasks, which should not be learned through the memorization of a large number of examples but through finding/learning the correct algorithm. \new{We hope these results will inspire practitioners to explore new model architectures when encountering catastrophic forgetting with standard models.}

\section*{Acknowledgements}

This work was motivated by a talk given by Irina Rish as part of ICML2021. We thank Rahul Chacko, Farshid Jafarpour, Stefano Sarao Mannelli, Andrew Saxe, Markus Spitzer and Nachi Stern for stimulating discussions. We have used Google JAX\cite{jax2018github} to perform our experiments. We acknowledge support from the CONEX-Plus programme funded by Universidad Carlos III de Madrid and the European Union's Horizon 2020 research and innovation programme under the Marie Sklodowska-Curie grant agreement No. 801538.









\bibliography{bib.bib}

\end{document}